\documentclass[runningheads]{llncs}
\usepackage[T1]{fontenc}
\usepackage{ulem}
\usepackage{color}
\usepackage{marvosym}
\usepackage{graphicx}
\usepackage{multirow}
\usepackage{booktabs}
\usepackage[table,xdraw]{xcolor}
\usepackage[colorlinks,linkcolor=blue]{hyperref}
\begin{document}
\title{Personalized Diagnostic Tool for Thyroid Cancer Classification using Multi-view Ultrasound}

\titlerunning{Personalized Diagnostic Tool for Thyroid Cancer in Multi-view US}
%
\author{Han Huang\inst{1,2,3} \and Yijie Dong\inst{4} \and Xiaohong Jia\inst{4} \and Jianqiao Zhou\inst{4} \and Dong Ni\inst{1,2,3} \and Jun Cheng\inst{1,2,3}\textsuperscript{(\Letter)} \and Ruobing Huang\inst{1,2,3}\textsuperscript{(\Letter)}}

\authorrunning{H. Huang et al.}
%
\institute{
\textsuperscript{$1$}National-Regional Key Technology Engineering Laboratory for Medical Ultrasound, School of Biomedical Engineering, Health Science Center, Shenzhen University, China\\
\email{ruobing.huang@szu.edu.cn; chengjun583@qq.com} \\
\textsuperscript{$2$}Medical Ultrasound Image Computing (MUSIC) Lab, Shenzhen University, China\\
\textsuperscript{$3$}Marshall Laboratory of Biomedical Engineering, Shenzhen University, China\\
\textsuperscript{$4$}Department of Ultrasound Medicine, Ruijin Hospital, School of Medicine, Shanghai Jiaotong University, China\\}

\maketitle        
\begin{abstract}
Over the past decades, the incidence of thyroid cancer has been increasing globally. Accurate and early diagnosis allows timely treatment and helps to avoid over-diagnosis. Clinically, a nodule is commonly evaluated from both transverse and longitudinal views using thyroid ultrasound. However, the appearance of the thyroid gland and lesions can vary dramatically across individuals. Identifying key diagnostic information from both views requires specialized expertise. Furthermore, finding an optimal way to integrate multi-view information also relies on the experience of clinicians and adds further difficulty to accurate diagnosis.
To address these, we propose a personalized diagnostic tool that can customize its decision-making process for different patients. It consists of a multi-view classification module for feature extraction and a personalized weighting allocation network that generates optimal weighting for different views. It is also equipped with a self-supervised view-aware contrastive loss to further improve the model robustness towards different patient groups. Experimental results show that the proposed framework can better utilize multi-view information and outperform the competing methods.
\end{abstract}

\section{Introduction}
\begin{figure}
\includegraphics[width=\textwidth]{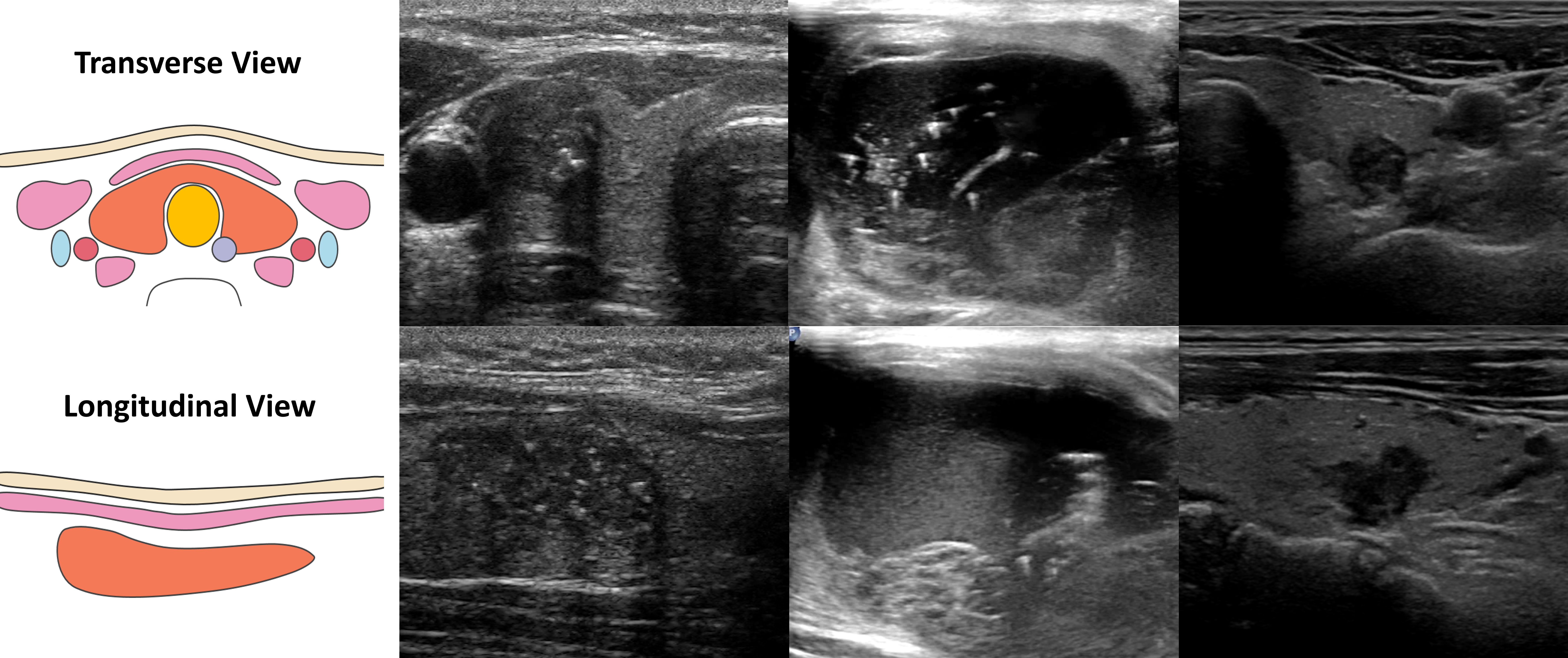}
\caption{Illustration of the transverse (upper row) and longitudinal (lower row) planes of the thyroid. Each row corresponds to the US images of one patient, while the anatomical diagrams are shown in the left. It can be seen that both the imaging conditions and the lesion appearance vary dramatically across individuals.
} \label{fig:cases}
\end{figure}

The world has witnessed a rapid global increase in the incidence of thyroid carcinoma during the past decades~\cite{cabanillas2016thyroid}. Accurate diagnosis allows access to timely intervention and helps to prevent over-diagnosis.
Ultrasound (US) is the primary technique used in thyroid cancer screening due to its wide availability, low cost and non-ionizing nature~\cite{hegedus2004thyroid}.
The correct interpretation of thyroid US requires spatial understanding of the anatomy (see Fig.\ref{fig:cases}) and accurate recognition of diagnostic related features. These necessitate both expertise and a high-level of experience, and may elude the less-experienced clinicians. 

To alleviate this issue, many computer-aided diagnosis (CAD) tools were proposed~\cite{chi2017thyroid,liu2017classification,zhao2022local}. 
For example, Chi et al.~\cite{chi2017thyroid} used a GoogLeNet to extract features of US and then applied a Random Forest classifier to identify thyroid nodules. 
Liu et al.~\cite{liu2017classification} combined deep features from CNN with conventional ones, forming a hybrid model for thyroid cancer classification.
Zhao et al.~\cite{zhao2022local} proposed a local and global features disentangled network to classify thyroid nodules into benign and malignant cases. These methods shed light on how to approach the problem, while they only focused on the primary transverse view of the US scans and neglected the other.
However, US scanning for thyroid is often done both in transverse and longitudinal planes in clinical practice~\cite{chaudhary2013thyroid}.
Other clinical studies also revealed that longitudinal planes help to promote diagnosis accuracy by excluding suspicious benign nodules and discovering potential malignancy markers that might not be visible in the transverse planes~\cite{sipos2009advances,moon2011taller}. 
For example, the column 2 of Fig.~\ref{fig:cases} shows an example whose longitudinal view indicates existence of micro-calcification, while its transverse view fails to provide this information. Similarly, the column 4 shows another case where the lesion seems to have a regular shape when viewed from the transverse direction, while its longitudinal direction exhibits the opposite information.
New diagnostic tool is needed to incorporate this rich multi-view information. 

Combining information from different views has been investigated previously to analyze other medical images. Modern approaches mainly leveraged deep learning (DL) models to extract and fuse multi-view information.
For example, Kyono et al.~\cite{kyono2019multi} presented a novel multi-view multi-task (MVMT) CNN model to fuse the four views of mammograms for breast cancer diagnosis. Pi et al.~\cite{pi2020automated} developed an attention-augmented deep neural network (AADNN) to detect bone metastasis using anterior and posterior views of X-ray images. Wang et al.~\cite{wang2021automated} used depth-wise separable convolution-based CNNs to organize the five views of echocardiograms. 
Note that, to the best of our knowledge, there also exist two multi-view studies in thyroid US~\cite{liu2019automated,chen2021computer}. The former~\cite{liu2019automated} essentially used a three-branch CNN model to classify thyroid nodules with different resolutions, while the latter~\cite{chen2021computer} referred to the deep features, statistical and texture features of US images as different "views" of thyroid lesions and combined this information through an ensemble model. Neither of them utilized both the transverse and longitudinal views of thyroid US images for lesion classification.

Despite showing promising results, these methods treated different views equally, overlooking the fact that one of the views might contain critical diagnostic information while the other should be de-emphasized. More importantly, as the lesions vary in size, shape, appearance across individuals (see Fig.\ref{fig:cases}), this disparity among the importance of different views should be customized to the particular condition of each patient. 
To address this, we propose a personalized diagnostic tool for automated thyroid cancer classification. Our contribution is three-fold: 
\begin{itemize}
    \item The first DL-based framework that intelligently integrates both the transverse and longitudinal views for thyroid cancer diagnosis using US.
    \item A personalized weighting allocation network that customizes the multi-view weighting for different patients. It can be trained end-to-end and does not require additional supervision.
    \item A self-supervised view-aware contrastive loss that considers intra-class variation inside patient groups and can further improve the model performance.
\end{itemize}
Experiment results showed that the trained model is able to intelligently exploit multi-view information and outperform state-of-the-art approaches in thyroid cancer diagnosis.

\section{Methods}
\begin{figure}[htb]
\centering
\includegraphics[width=0.98\textwidth]{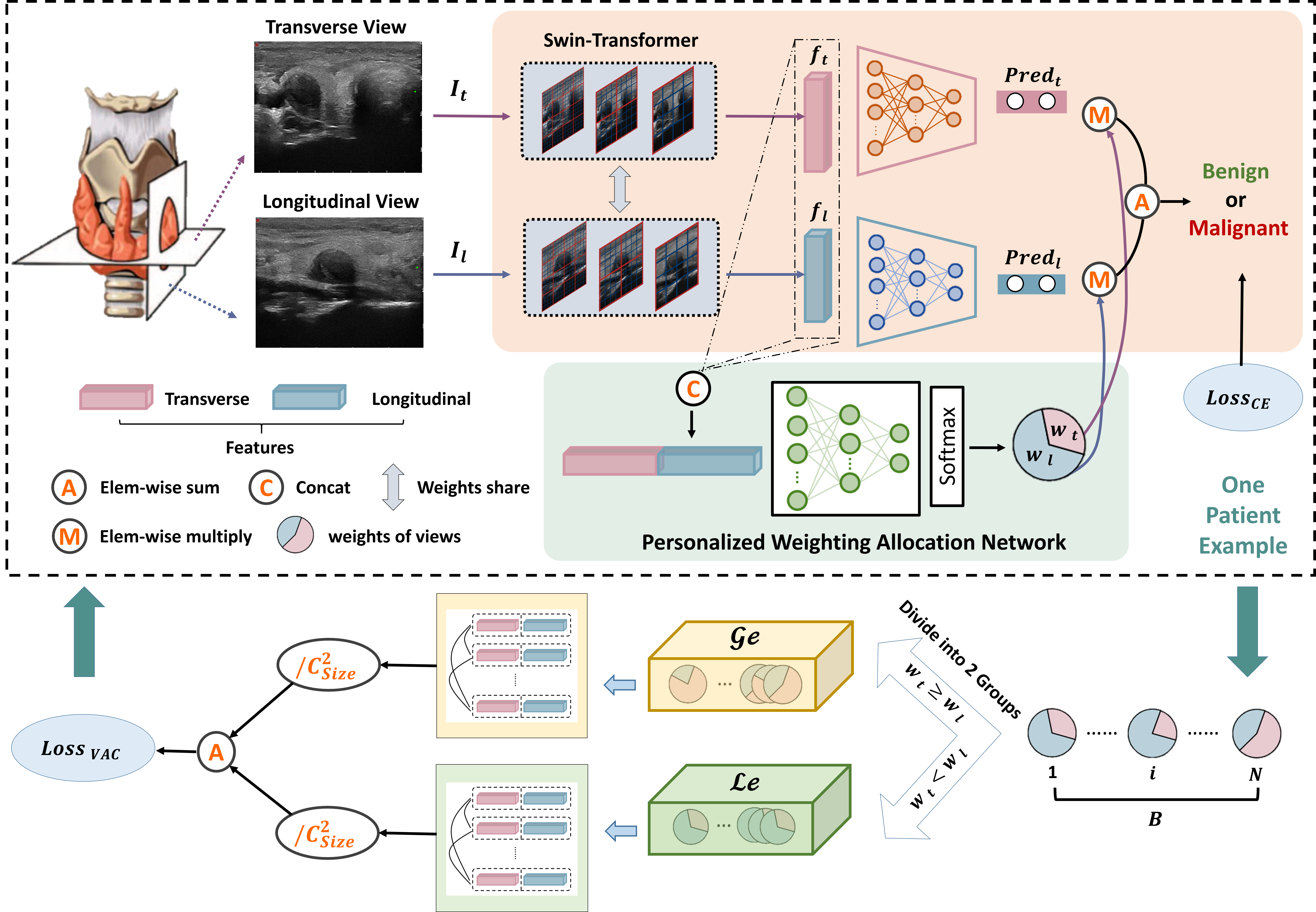}
\caption{Schematic of the overall framework. Multi-view thyroid US images are first processed by the swin-transformer backbone (grey block). The derived multi-level features are sent to subsequent MLPs to generate view-specific predictions, respectively. The PAWN then automatically assigns the optimal multi-view weighting for each patient ($w_t,w_l$). These pairs of weights are also collected to separate samples and the view-aware contrastive loss $\mathcal{L}_{VAC}$ is calculated.} \label{fig:overallframwork}
\end{figure}

Processing information from heterogeneous sources is a non-trivial task, while combining them according to individual specifications can be more challenging and is rarely explored. To address this, we propose a framework that learns to assign customized multi-view weighting for thyroid cancer diagnosis. Fig.~\ref{fig:overallframwork} displays the overall framework. The inputs are first passed to a multi-view classification network (MVC) for feature extraction and pre-prediction. Then, the features are sent to a novel personalized weighting allocation network (PAWN) for multi-view weighting generation. The training procedure is also equipped with a view-aware contrastive loss (VACL) (lower part of Fig.~\ref{fig:overallframwork}). Details of each component is explained next.

\subsection{Multi-view Classification Model}
Extracting multi-view features automatically while dealing with view-related information efficiently has always been a challenge in multi-view learning.
The proposed MVC model uses a two-branch network to address this, where each branch corresponds to one anatomical view of thyroid US (orange block in Fig.\ref{fig:overallframwork}). The swin-transformer architecture is adopted as the feature extraction backbone due to its prominent modeling capacity~\cite{liu2021swin}. Note that two branches share the same parameters to reduce over-parameterization. This design can help prevent overfitting and also encourage interaction between different views. 
Formally, the MVC model accepts a pair of thyroid US images: $I_t^p$ and $I_l^p$ of the same patient $p$. The former is collected from the transverse direction, while the latter corresponds to the longitudinal view. The feature extraction backbone then yields corresponding high-level features: $f_t^p=ST(I_t^p),f_l^p=ST(I_l^p)$. Then, $f_t^p,f_l^p$ are fed into unique multilayer perceptron (MLP) heads respectively to generate view-specific decisions, denoted as $pred_t^p$ and $pred_l^p$. This structure boosts the model's flexibility and allows learning view-specific representations and making view-independent decisions. The $f_t^p,f_l^p$ are sent to the subsequent PAWN, while $pred_t^p$ and $pred_l^p$ are saved to generate the final prediction result.   

Note that the MVC network is a self-contained model and can be trained independently to warm up the learning. Specifically, the output of MVC model is formulated as $\hat{pred^p}= 0.5*pred_t^p + 0.5*pred_l^p$, and evaluated against the ground truth. The model can then be optimized through a standard cross-entropy loss ($\mathcal{L}_{CE}$) for a few epochs. The PWAN is then attached to the pre-trained MVC for end-to-end optimization. Empirical experiments found that synchronized optimization facilitates the learning of holistic features.

\subsection{Personalized Weighting Generation} \label{sec:PWAN}
An ideal multi-view model should make predictions using all the available information but may put emphasis on a particular view based on the specific situation. To fulfill this, a lightweight network named PAWN was designed to generate the optimal weighting for different views to yield accurate diagnosis. Instead of directly processing raw images, the PAWN operates on high-level features $(f_t^p,f_l^p)$ of different views. This helps to reduce repetitive computation and can accelerate convergence. Moreover, it also forces the PAWN to allocate weights explicitly based on diagnosis-related features and coordinates parameter updates across modules. The PAWN, therefore, only requires a simple architecture to operate, i.e., a three-layer MLP (with a hidden size of 256, 128, 32, respectively). It generates optimal weighting for each view which can be denoted as: $(w_t^p,w_l^p)=Softmax(PWAN(f_t^p,f_l^p)),\,s.t.\,w_t^p+w_l^p=1$. The $Softmax()$ operation enforces the subjective constraint. The value of $(w_t^p,w_l^p)$ controls the impact of the corresponding view in the final decision-making process. Given view-independent predictions $(pred_t^p,pred_l^p)$, the final classification result for patient $p$ is calculated as:
\begin{equation}
\hat{pred^p}= w_t^p*pred_t^p + w_l^p*pred_l^p.
\end{equation}
As the whole process is differentiable, the parameters of PWAN can then be updated using back-propagation given the ground truth classification label. This is especially beneficial as manually assigning multi-view weights is infeasible and might suffer from observer subjectivity. During the inference, the operator can inspect both the view-independent decision $pred_t^p$, $pred_l^p$ and the combined one $\hat{pred^p}$. This might also help to strengthen the interpretability of the model decision-making process. 

Note that similar to the state-of-the-art subject-customized approaches \linebreak (e.g.\cite{panda2021adamml}), it is also possible to adopt a view selected strategy to generate the final outcome. However, this approach discards part of the inputs and inevitably loses information. We argue that all the views should be considered jointly for accurate thyroid cancer diagnosis and utilized in a weighted way. 
Later validation experiments also validate this conjecture (see Section.4). 

\subsection{View-aware Contrastive Loss}
Learning effective visual representations without human supervision is a long-standing problem. Recently, discriminative approaches based on contrastive learning in the latent space have shown great promise~\cite{khosla2020supervised,hjelm2018learning,chen2020simple}. 
The core of contrastive learning is to penalize dis-similar features in high-level space to increase model robustness and may help to alleviate over-fitting. As lesions have large variations in size, shape, appearance, and more importantly, different diagnostic features in different anatomical planes, rashly restricting all representations may produce adverse effects and confuse the training. Therefore, we argue that applying contrastive learning to multi-view data requires further consideration. In specific, we propose a novel contrastive loss equipped with view awareness, which segregates samples based on whether they emphasize the same view in lesion classification and only constrain similarity within the same group.

\sloppy Formally, given a training batch $B=\{1,...,i,...,N\}$, the corresponding feature set and the weight set can be defined as: $F=\{(f_t^1,f_l^1),...,(f_t^N,f_l^N)\}$, $W=\{(w_t^1,w_l^1),...,(w_t^N, w_l^N)\}$, respectively. $B$ can be divided into two groups:  $Ge=\{i|i \in B, w_t^i \ge w_l^i\}$, and $Le=\{i|i \in B, w_t^i < w_l^i\}$, while $Ge + Le = B $. It is clear to see that cases belonging to the same group have similar weighting distributions over different views and their features should be relatively close in high dimensional space. The VACL can be defined as follows: 
\begin{equation}
    \mathcal{L}_{VAC}=\frac{\sum_{i,j \in Ge \atop i\not=j} Dis(i,j)}{C_{Size(Ge)}^2 +\epsilon} +\frac{\sum_{i,j \in Le \atop i\not=j} Dis(i,j)}{C_{Size(Le)}^2 +\epsilon}
\end{equation}
where $Dis(i,j) = \Arrowvert f_t^i - f_t^j \Arrowvert_1 + \Arrowvert f_l^i - f_l^j\Arrowvert_1$ measures the mean absolute distance between the high-level features of the patient $i$ and $j$. $C_{Size}^2$ represents the combination formula and is used to normalize the numerator based on total number of calculations. $Size(Ge), Size(Le)$ represent the number of samples in each group. $\epsilon$ is a small positive number to avoid division by 0. $\mathcal{L}_{VAC}$ can therefore selectively encourage feature resemblance between samples with similar diagnostic patterns and avoid penalizing the rest.
Finally, the overall loss function can be summarized as follows:
\begin{equation}
\mathcal{L}_{total} = \mathcal{L}_{CE} +\lambda * \mathcal{L}_{VAC},
\end{equation}
where $\lambda$ is a hyper-parameter that controls the influence of the proposed contrastive constraint. 

\section{Materials and Experiments}
We validate our approach on an in-house dataset containing 4529 sets of multi-view US images of thyroid nodules. Each set contains a pair of multi-view (a transverse and a longitudinal views) US images collected from one patient. This study was approved by the local Institutional Review Board. We randomly split the dataset at the patient level into 7:1:2 for training, validation, and test. All images are resized to 224 $\times$ 224. Biopsies were carried out for each patient used as the ground truth.

We compare the performance of the proposed model with that of single-view models and popular multi-view approaches (i.e., MVMT~\cite{kyono2019multi}, AADNN~\cite{pi2020automated}). The single view models use the same feature extraction backbone as the proposed model but are trained only using transverse (row 1, Tab.~\ref{tab:results}) or longitudinal view (row 2, Tab.~\ref{tab:results}). Note that some multi-modal models are closely related to this task as well. We therefore select two state-of-the-art works: AW3M~\cite{huang2021aw3m} and AdaMML~\cite{panda2021adamml}, as the former treats different branches differently, while the latter selects different modalities to perform classification for different patients. For ablations study, we also implement the proposed model without the PAWN and VACL (row 7, Tab.~\ref{tab:results}) and only without the VACL (row 8, Tab.~\ref{tab:results}) to verify the effectiveness of each component of the proposed model. Furthermore, we also switch the VACL with an original contrastive loss~\cite{khosla2020supervised} (row 9, Tab.~\ref{tab:results}) to examine whether the view-dependent design of VACL is effective.

Different augmentation strategies are applied for all experiments, including scaling, rotation, flipping, and mixup. Weights pre-trained from ImageNet are used for initializing. We use the AdamW optimizer with a learning rate of 5e-4. The weight decay is set to 0.05. $\epsilon$ is set to 0.001. The value of $\lambda$ is defined by considering the order of magnitude of the two losses and is empirically set to be 0.01. The MVC network is trained for 300 epochs together with the PWAN.
Accuracy (ACC), sensitivity (SEN), precision (PRE), specificity (SPE) and F1-score are used as evaluation metrics. All experiments are implemented in PyTorch with a NVIDIA RTX A6000 GPU. 

\section{Results and Discussion}
\begin{table}
\centering
\caption{Results of the comparison experiments. Each row represents one approach while each column corresponds to different evaluation metrics.}\label{tab:results}
\begin{tabular}{c c c c c c c}
\toprule
 & Methods & ACC(\%) & SEN(\%) & SPE(\%) & PRE(\%) & F1-score(\%)\\
\hline
\multirow{2}*{Single-view}& Transverse & 79.21 & 85.57 & 66.85 & 83.38 & 84.46\\
 & Longitudinal & 79.58 & 85.29 & 68.49 & 84.03 & 84.65\\
\hline
\multirow{6}*{Multi-view}& MVMT~\cite{kyono2019multi} & 81.00 & 85.38 & \textbf{72.50} & \textbf{85.78} & 85.58\\
 & AADNN~\cite{pi2020automated} & 80.75 & 82.75 & 68.12 & 84.18 & 85.69\\
 & AW3M~\cite{huang2021aw3m} & 80.45 & 90.07 & 61.75 & 82.07 & 85.88\\
 & AdaMML~\cite{panda2021adamml} & 65.90 & 65.14 & 67.40 & 79.52 & 71.61\\
 & Ours w/o PAWN & 81.87 & 89.41 & 67.21 & 84.13 & 86.69\\
 & Ours w/o VACL & 82.18 & \textbf{90.72} & 65.57 & 83.66 & 87.05\\
 & Ours with OCL & 80.94 & 89.03 & 65.21 & 83.26 & 86.05\\
 & Ours & \textbf{83.29} & 89.97 & 70.31 & 85.49 & \textbf{87.67}\\
\bottomrule
\end{tabular}
\end{table}
As shown in row 1 and 2 in Tab.~\ref{tab:results}, comparable performance was obtained for the two single-view models. This shows that both views provide informative features for cancer diagnosis. In fact, the longitudinal model scored slightly higher ACC and F1-score in our datasets, suggesting its necessity in decision making. On the other hand, most multi-view models exhibited better performance than that of single-view ones. This indicates that there exists complementary information hidden in different views of thyroid US, and their combination could lead to a more accurate diagnosis. Compared with AADNN, the MVMT obtained higher ACC and PRE. It might be caused by its larger size with greater modeling capacity.
The AW3M model scored higher F1-score, displaying a balanced performance. This may stem from the fact that this model breaks the equilibrium of different branches and learns to weight different views differently. However, it overlooks that this weighting should vary according to individual specification of each patient. As a result, it performed inferiorly than the proposed method, which scored 87.67\% in F1-score (last row in Tab.\ref{tab:results}). Another interesting comparison method is the AdaMML model, the only approach that customizes its decision-making for different patients. Nevertheless, it exhibited inferior performance in our dataset (row 6 in Tab.\ref{tab:results}). This may result from the fact that this approach was originally proposed to analyze multi-modal data for video classification and learns to discard modalities to improve accuracy. This approach could handle data redundancy and noise but inevitably loses some information as well. However, in our task, both views contain crucial information and should be jointly considered during classification. On the contrary, our personalized diagnostic tool avoids information loss by allocating different multi-view weights for different patients. It scored ACC=83.29\%, SEN=89.97\%, SPE=70.31\%, PRE=85.49\% and F1-score=87.67\%, and outperformed all competing methods, indicating that it is a suitable solution for our task.

As an ablation study, we investigated how the proposed framework would perform without the PWAN and VACL. It can be observed in Tab.~\ref{tab:results} that the MVC model alone could achieve promising performance (row 7), proving itself as an effective baseline for multi-view classification. Meanwhile, the introduction of PAWN boosted the model performance (row 8) as this design enables personalized weighting for different views. 
Moreover, the addition of VACL continued to improve the model's accuracy (compare row 8 with 10). We conjecture that this constraint helps to harness the learning of high-level representations and increases the robustness of the model. Note that as both the design of PAWN and VACL is general, they could be easily extended to other multi-view or multi-modal problems in future applications. 
To fully investigate the efficacy of the VACL, we also implemented the same framework with an original contrastive loss (row 9). Results show that this variant scored inferiorly than that trained with the proposed VACL (row 10). This suggests that segregating samples based on their importance of views during contrastive learning is more beneficial than the vanilla one. 
Comparing row 9 and 8, the additional contrastive loss led to a decrease of 1.69\% in SEN, 1.24\% in ACC and 1\% in F1-score. It further indicates that blindly enforcing features resemblance might be detrimental for the challenging thyroid cancer diagnosis.

\section{Conclusions}
In this paper, we proposed a personalized diagnostic tool for thyroid cancer diagnosis using multi-view US. 
Its design encourages interactions between different views via a weight-sharing multi-branch backbone, while also allows asymmetric emphasis on different views through a portable weighting allocation network. It also leverages a novel view-aware contrastive loss to further increase the model robustness.
The experimental results showed that the proposed method outperformed single-view models and other state-of-the-art multi-view approaches. 
The design of this framework is general, and could be applied to other multi-view applications.

\subsubsection{Acknowledgements}
This work was supported by the National Natural Science Foundation of China (No. 62101342, 62171290, and 61901275); Shenzhen-Hong Kong Joint Research Program (No. SGDX20201103095613036), Shenzhen Science and technology research and Development Fund for Sustainable development project (No. KCXFZ20201221173613036); National Natural Science Foundation of China (No. 82071928); Shenzhen University Startup Fund (2019131).

\bibliographystyle{splncs04}
\bibliography{references}

\end{document}